%% file: ICML2024_conference.tex
\documentclass{article}

\usepackage{microtype}
\usepackage{graphicx}
\usepackage{subfigure}
\usepackage{booktabs} 

\usepackage[table]{xcolor}
\usepackage{color}

\usepackage{algorithm}
\usepackage{algpseudocode}

\usepackage{hyperref}
\usepackage{url}
\usepackage[T1]{fontenc}
\usepackage[utf8]{inputenc}
\usepackage{babel}
\usepackage[font=small,labelfont=bf]{caption}

\usepackage{microtype}
\usepackage{graphicx}
\usepackage{subfigure}
\usepackage{booktabs} 
\usepackage{multirow}
\usepackage{soul}
\usepackage{mathtools}
\usepackage{amsmath,amsfonts,amsthm}
\usepackage{graphicx}
\usepackage{float}
\usepackage{lipsum}
\usepackage{textcomp}
\usepackage{xcolor}
\usepackage{breqn}
\usepackage{soul}
\usepackage{placeins}
\usepackage{hhline}
\usepackage{tabularx}
\usepackage{enumitem}
\usepackage{setspace}
\usepackage{booktabs}
\usepackage{adjustbox}
\usepackage[normalem]{ulem}
\usepackage{indentfirst}
\usepackage{titlesec}
\usepackage{wrapfig}

\usepackage{hyperref}

 \usepackage[accepted]{icml2024}

\usepackage{amsmath}
\usepackage{amssymb}
\usepackage{mathtools}
\usepackage{amsthm}

\usepackage[capitalize,noabbrev]{cleveref}

\theoremstyle{plain}

\theoremstyle{definition}

\theoremstyle{remark}

\usepackage[textsize=tiny]{todonotes}

\icmltitlerunning{Prompting-based Temporal Domain
Generalization}

\begin{document}

\twocolumn[
\icmltitle{Prompting-based Temporal Domain
Generalization}

\newcommand{\fix}{\marginpar{FIX}}
\newcommand{\new}{\marginpar{NEW}}


\icmlsetsymbol{equal}{*}

\begin{icmlauthorlist}
\icmlauthor{Sepidehsadat Hosseini}{yyy,comp}
\icmlauthor{Mengyao Zhai}{comp}
\icmlauthor{Hossein Hajimirsadeghi}{comp}
\icmlauthor{Frederick Tung}{comp}


\end{icmlauthorlist}

\icmlaffiliation{yyy}{Department of Computing Science, Simon Fraser University, Canada.}
\icmlaffiliation{comp}{Borealis AI, Canada  }

\icmlcorrespondingauthor{Sepidehsadat Hosseini}{sepid.hosseini@borealisai.com}

\icmlkeywords{Machine Learning, ICML}

\vskip 0.3in
]



\printAffiliationsAndNotice{\icmlEqualContribution} 

\input{sections/abstract}
\input{sections/intro_icml}

\input{sections/related_work}

\begin{figure*}[t!]
    \centering    \includegraphics[width=\textwidth]{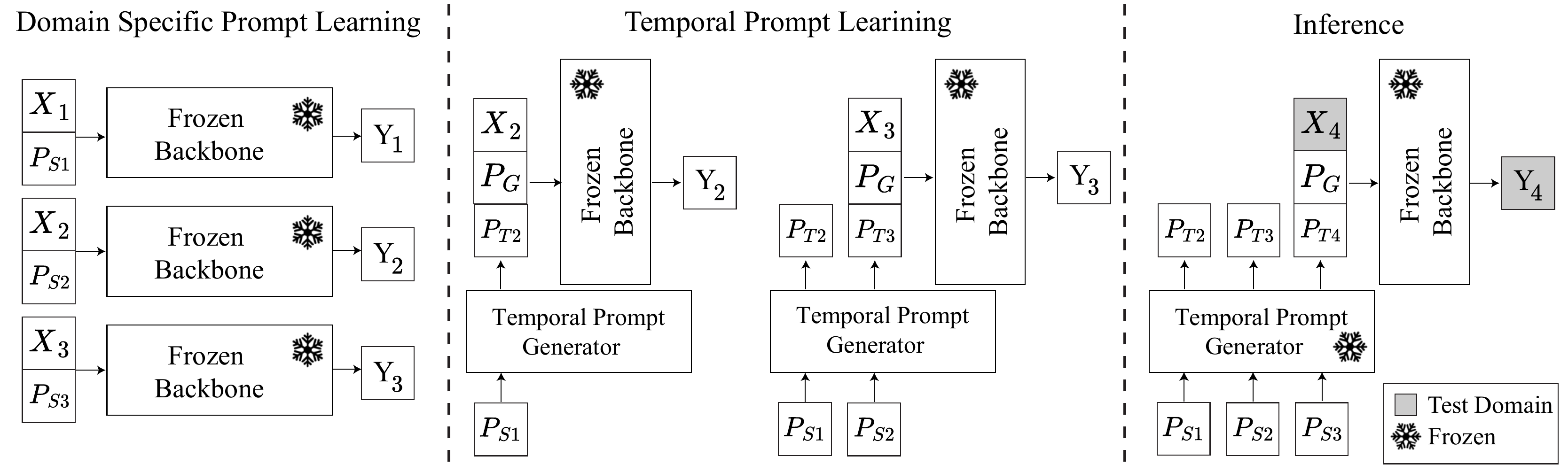}
    \caption{Illustration of the proposed method with 
    a set of source domains $D_1$, $D_2$, and $D_3$ given during training and a target (unobserved future) domain $D_4$ used only during testing.
    First, a backbone network is trained on the combined source domains in a pre-training phase. Then, domain-specific prompts \(P_{S1}, P_{S2}, P_{S3}\) are learned independently on each source domain (while keeping the backbone network frozen) to learn the characteristics of each indexed domain separately. Next, a temporal prompt generator is trained to transform the domain-specific prompts to temporal prompts \((P_{T2}, P_{T3}, P_{T4})\) which can capture the temporal dynamics and concept drifts within the sequence of domains. Finally, to capture the general knowledge across all domains, the general prompts \(P_G\) are learned. During inference, the combination \([P_{T4}; P_G, X_4]\) is fed to the frozen backbone to perform the task on the target domain \(D_4\).
    }
    \label{fig:methodl}
\end{figure*}

\input{sections/method}

\input{sections/experiments}

\input{sections/conclusion_icml}


\bibliography{iclr2024_conference}
\bibliographystyle{icml2024}

\newpage
\appendix

\input{sections/appendix}


\end{document}

%% file: sections/abstract.tex
\begin{abstract}
Machine learning traditionally assumes that the training and testing data are distributed independently and identically. However, in many real-world settings, the data distribution can shift over time, leading to poor generalization of trained models in future time periods. 
This paper presents a novel prompting-based approach to temporal domain generalization that is parameter-efficient, time-efficient, and does not require access to 
future data
during training. Our method adapts a 
trained model to temporal drift by learning global prompts, domain-specific prompts, and drift-aware prompts that capture underlying temporal dynamics. 
Experiments on classification, regression, and time series forecasting tasks demonstrate the generality of the proposed approach.
The code repository will be publicly shared.
\end{abstract}

%% file: sections/intro_icml.tex
\section{Introduction}

\begin{figure}[t]
\label{fig:teaser}
\includegraphics[width=\linewidth]{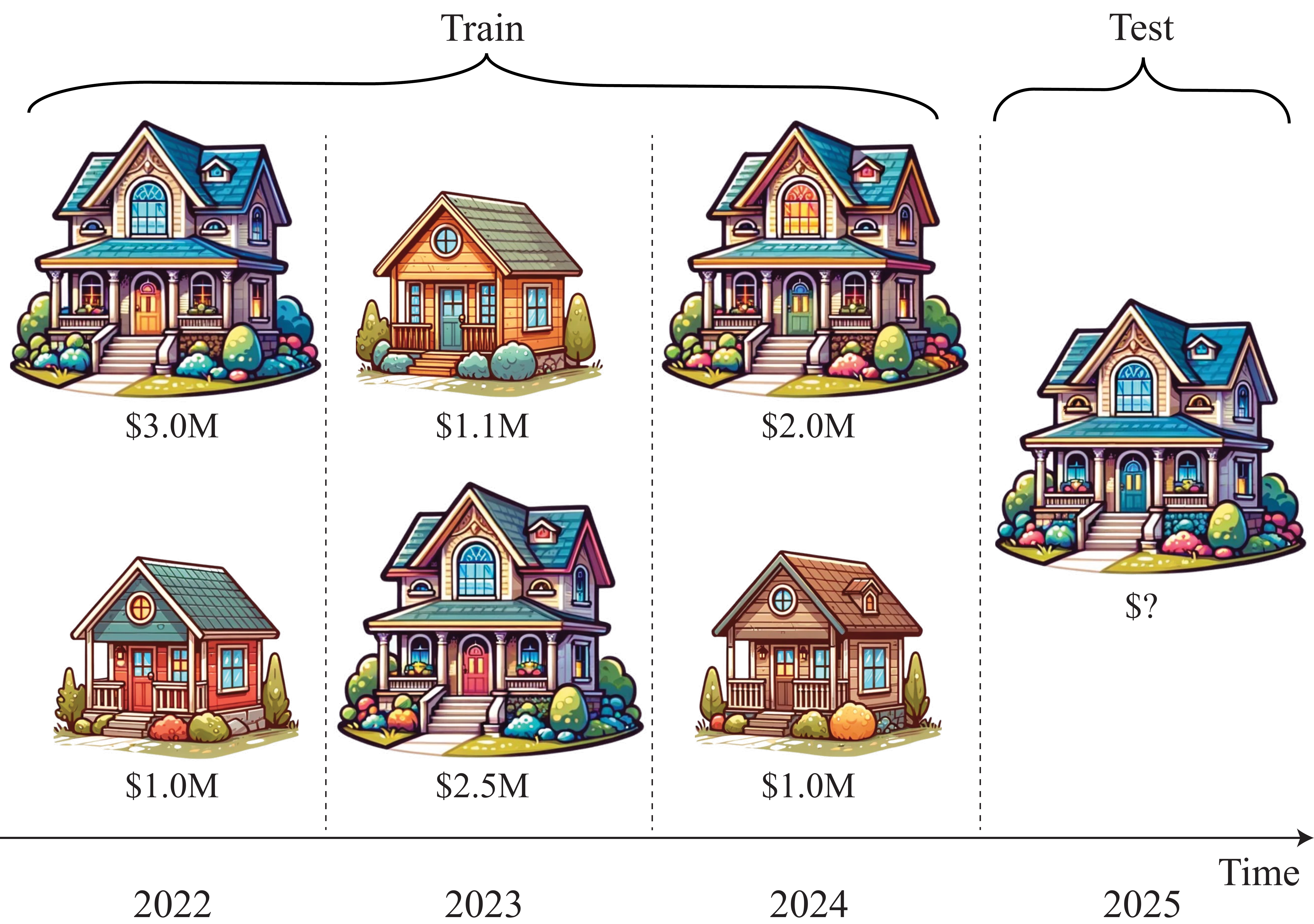}
\caption{Machine learning models trained under i.i.d. assumptions can struggle to generalize to the future. Prompting-based temporal domain generalization offers a general, parameter efficient, and time efficient way to adapt trained models to future time periods, without access to future data. In this toy example (see Introduction), 
an adapted model may anticipate that three-storey, many-bedroom houses with similar features will be priced \$0.5M less next year (2025) than at present. House images generated by DALL-E.}
\centering
\end{figure}

Most machine learning algorithms rely on the assumption that the training and testing data are independently and identically distributed (i.i.d.). However, real-world settings are often non-i.i.d. Distribution shift and concept drift over time can make the learning problem more challenging and lead to poor generalization of trained models in future time periods. The unobservability of data from the future further adds to the difficulty of making trained models more ``future-proof".

Domain generalization (DG) algorithms address the learning problem where we need to adapt models trained on a source domain to a novel target domain without the availability of the target data, labelled or unlabelled~\cite{yue2019domain,prakash2019structured,shankar2018generalizing,volpi2018generalizing,hu2021domain,triantafillou2021learning,kim2021self,wang2020meta}. 
By casting different time periods as different domains, we can draw on the rich DG literature to find principled ways to discover general representations among different domains and learn domain-invariant features. This takes us \emph{part} of the way towards our goal of future-proofing a trained model. The remaining essential ingredient is a way to capture the temporal dynamics of the data -- how the data distribution evolves over time -- which allows us to extrapolate the model to unobserved future temporal domains. 

As an example, consider the task of predicting house prices given information about the property's physical characteristics, such as square footage, number of bedrooms, number of bathrooms, and location. Since house prices are influenced by macroeconomic conditions and demographic trends that change over time, a regression model trained on data collected from the past few years could have poor predictive power next year~\cite{yin2022spatial}. However, suppose that the macroeconomic and demographic factors change gradually over time. In that case, we can extrapolate their influence into the short-term future, and adapt the regression model to make more accurate predictions. 
For example, suppose that the population in your country has been steadily aging over the past several years, which reduces the overall demand for many-bedroom houses. A future-robust model would anticipate that the demand will continue to fall for many-bedroom houses and adapt the price predictions for these houses accordingly: given the same features, a many-bedroom house next year will be priced some amount less than this year. 

In this paper, we contribute a new algorithm for \emph{temporal domain generalization}~\cite{bai2023temporal, nasery2021training} that is parameter and time efficient. Our solution is inspired by \emph{prompting}: an efficient adaptation mechanism popularized by large language models (LLMs). In LLMs, a prompt is a piece of natural text that is appended to the user input for the purposes of conditioning or adapting the trained model's text generation process. More generally, a prompt can be any input (e.g. a learnable vector) that is appended to the network input for conditioning and adapting the trained model's output generation process. 
Our approach learns a set of global prompts, domain-specific prompts, and temporal (drift-aware) prompts. The drift-aware prompts capture the temporal changes in the domain-specific prompts and forecast the prompts that will be effective in subsequent time periods. All prompts are generated as real-valued vectors. The generality of our approach allows it to be easily applied to a wide range of network architectures and prediction tasks.

To sum up, we propose the first prompting-based temporal domain generalization method for adapting trained models to data distribution shifts over time -- a challenging, non-i.i.d. problem setting that is common in real-world settings. The benefits offered by prompting-based temporal DG include: 
\begin{itemize}
\item \textbf{Parameter and time efficiency.} Only a small number of parameters shared across all domains are needed for prompt generation, and no additional parameters are needed for the target (future) domain. Observations from the future domain (e.g. next year's house prices) are not required during training.
\item \textbf{Generality.} Prompting-based temporal DG supports many applications, including classification, regression, and time series forecasting, as illustrated in the experimental results.
\end{itemize}

%% file: sections/related_work.tex
\section{Related Work}
\label{sec:related}

\textbf{Domain generalization and adaptation} are research fields that have garnered significant attention in recent years due to their practical significance in real-world applications~\cite{Ganin15,tzeng2017adversarial,tremblay2018training,shankar2018generalizing,volpi2018generalizing,zhou2020deep}. The primary goal of domain adaptation (DA) is to tailor models to specific target domains, using the similarities that exist between these domains~\cite{ben2010theory,wang2018deep}. Continuous domain adaptation, a subset of DA, addresses the adaptation to domains characterized by continuous variables~\cite{hoffman2014continuous,ortiz2019cdot,lao2020continuous,wang2020continuously,yang2016multivariate}. This may include temporal domain adaptation, which deals with domains that evolve over time. For instance,~\cite{courty2016optimal,gong2012geodesic} adapted their training loss to account for future data derived from prior domains. Similarly, the method proposed by~\cite{mancini2019adagraph,scaleformer} involves time-sensitive deep neural network parameters to control their evolution over time. Their network possesses domain-specific and domain-generic parameters, with the former integrating an added constraint that considers the similarity between domains. Meanwhile, other approaches like~\cite{wang2020continuously, ganin2016domain} focus on learning time-invariant representations using adversarial methods.

Domain generalization (DG) methods build upon the insights from domain adaptation and aim to enhance the generalization capability of models across \emph{unseen} target domains, where the data distribution may differ significantly from the source domain. 
DG techniques encompass a range of strategies, as outlined in~\cite{wang2021generalizing}. DG methods can be categorized into three groups based on their focus: first, data manipulation methods, which include data augmentation by manipulating input data through domain randomization~\cite{yue2019domain,prakash2019structured}, adversarial data augmentation~\cite{shankar2018generalizing,volpi2018generalizing,nazari2020domain,khirodkar2019domain} and data generation~\cite{qiao2020learning,liu2018unified,zhao2020learning,garg2020learn}; second, representation learning by either applying domain-invariant representation learning techniques~\cite{qi2021unsupervised,fan2021adversarially,mitrovic2021representation} or feature disentanglement techniques~\cite{hu2021domain,triantafillou2021learning,nam2021reducing,sun2021recovering} to improve generalization; third, learning strategy methods that exploit various strategies such as ensemble learning~\cite{wu2021collaborative,dubey2021adaptive}, meta-learning~\cite{kim2021self,wang2020meta}, and gradient operations~\cite{tian2021neuron,rame2021fishr} to enhance the overall generalization capability.

Traditional DG methods are designed for categorical domains. 
Temporal DG is a nascent area that addresses ongoing changes in the data distribution over time, referred to as concept drift. 
Unlike standard DG, which aims for generalized representations across different domains, temporal DG focuses on capturing the 
temporal dynamics of the data, enabling generalization to unseen future temporal domains. 
The Gradient Interpolation (GI)~\cite{nasery2021training} method uses adversarial training to generalize over time, altering the leaky ReLU activation for time dependence. 
The state-of-the-art temporal DG approach, Drift-Aware Dynamic Neural Networks 
(DRAIN)~\cite{bai2023temporal}, 
captures how data distributions and network weights evolve over time and predicts effective network weights for future domains. 
Instead of generating an entire set of network weights, our approach generates a set of \emph{prompts}: learnable vectors that are appended to the network input. We show in the experiments how this provides a significantly more parameter-efficient solution, making temporal DG accessible to larger state-of-the-art architectures such as Transformers, without loss in generalization performance.

\mysubsubsection{Prompting Mechanism} The concept of prompt-based learning has gained significant traction in the field of natural language processing (NLP) for adapting pre-trained language models (PLMs) to various downstream tasks. This framework involves conditioning the model with additional instructions to perform specific tasks. Elmo \cite{peters-etal-2018-deep}, Bert \cite{devlin2018bert}, and~\cite{brown_2020_gpt3} introduced the approach of fine-tuning PLMs for downstream tasks through fixed prompting functions. This technique has succeeded particularly in few-shot classification tasks like sentiment analysis and natural language inference \cite{gao-etal-2021-making,liu2021gpt}, where manually designed prompts were employed.

However, formulating a prompting function is challenging and often demands heuristic knowledge. In response to this challenge, recent efforts such as soft prompts \cite{lester2021, vu2021spot, gu2021ppt}, P-tuning V2 \cite{liu2021pre}, and prefix tuning \cite{li2021prefix} have 
proposed
to treat prompts as adaptable parameters. It is worth noting that prompts encapsulate task-specific supervision with notably fewer supplementary parameters than competing techniques, such as Adapter \cite{wang2020k,pfeiffer2020adapterfusion} and LoRA \cite{hu2021lora}.

 A different yet related angle to this topic is the casting of language modelling as a sequence-to-sequence task. This approach employs full transformer models, like the encoder-decoder paradigm, to autoregressively generate masked or altered token spans from input sequences \cite{raffel_2020_t5, lewis-etal-2020-bart}. The T5 model, introduced by~\citet{raffel_2020_t5}, exemplifies this concept by treating every task as generative, where tasks are prefixed with a specific phrase to denote the operation. This approach has spiked different exploration across numerous areas, from adapting language models for diverse utilities \cite{brown_2020_gpt3}, extracting sentiment or theme-centric details \cite{jiang2020can,sun2020conditioned, shin2020autoprompt, haviv2021bertese}, enhancing fine-tuning efficiencies~\cite{li2021prefix, scao2021}, to functioning as few-shot learning techniques \cite{gao-etal-2021-making, schick-schutze-2021-exploiting}. 

Moreover, researchers have studied the transferability of prompts \cite{wang-etal-2021-transprompt, vu2021spot,su2021transferability}, seeking to enhance the efficacy of prompt tuning across various tasks. Methods such as SPoT~\cite{vu2021spot} choose a prompt based on a similarity metric, whereas ATTEMPT~\cite{asai2022attempt} incorporates an attention mechanism to draw from source prompts, initializing the prompt for its designated task. \citet{wang2023multitask} achieved a universal prompt by decomposing and distilling knowledge from source prompts. 
However, none of these approaches are designed to handle temporal drift where the target domain is unseen.

%% file: sections/method.tex
\section{Method}

We address the problem of adapting a trained model to future time periods under the realistic setting where the data distribution evolves over time. Prompt tuning is a well-known mechanism for adapting a trained model to downstream tasks efficiently (see \cref{sec:related}). More specifically, prompt tuning comprises two main components: prompts and a frozen backbone transformer network. The prompts are prepended to the inputs, guiding the frozen transformer in adapting to different downstream tasks. Note that the backbone transformer is fixed and only the prompts are possible learnable parameters. 
In this section, we describe a new prompting-based temporal domain generalization method that uses prompt tuning to adapt a trained model to unseen future domains, without requiring data points from the target (unseen future) domains during training.

Our method utilizes a pre-trained model (Section~\ref{method:pretraining}) and two types of learnable prompts: domain-specific prompts $P_{S(t)}$ (Section~\ref{method:domain-specific-prompt}) and temporal prompts $P_{T(t)}$ (Section~\ref{method:temporal-prompt}). The domain-specific prompts encode the domain specific information, and the temporal prompts aim to capture the dynamics associated with temporal drift, and are generated using the domain-specific prompts. In Figure \ref{fig:methodl}, the left and middle subfigures illustrate the training procedure, and the right subfigure depicts inference.

\subsection{Backbone Network Pre-Training} 
\label{method:pretraining}

A key component of prompt-tuning is the frozen pre-trained backbone network, which should be general enough to lay a good foundation for later tasks. For temporal domain generalization, we pre-train a backbone network on the data combining all source domains. 

Denote a set of temporal domains by $\{D_t = (X_t, Y_t)\}$, where $\{D_t | 1 \leq t \leq \tau\}$ represents the source domains, $\{D_t | t > \tau \}$ represents the target (unseen future) domains, and \( X_{t} \) and \( Y_{t} \) are the inputs and outputs for domain $t$, respectively. A transformer-based network represented as \(f_\theta\) is pre-trained as the backbone model on $\{D_t | 1 \leq t \leq \tau\}$. This network is pre-trained on the combined datasets from all source domains and the goal is to train the \(f_\theta\) maximizing the likelihood $\mathcal{P}_\theta(Y_{1:\tau}| X_{1:\tau})$. After pre-training, the $f_\theta $ weights are fixed in all later steps.

\subsection{Domain-specific Prompt Learning} 
\label{method:domain-specific-prompt}

The backbone network in Section~\ref{method:pretraining} is pre-trained on the data aggregated across all source domains, without considering the differences in the individual domains. Intuitively, the pre-trained network captures ``average" or ``general" knowledge and can fail to capture details that are characteristic of particular domains. Therefore, we adopt prompts to capture domain-specific information. 
For each domain $D_t$, we prepend the input $X_t$ with a prompt $P_{S(t)} \in \mathbb{R}^n$, which are learnable parameters. The combined result, represented as $[P_{S(t)}; X_t]$, is then processed by the frozen backbone network \(f_\theta\). 
To learn prompt $P_{S(t)}$, the model is trained to maximize the likelihood $\mathcal{P}_\theta(Y_t | [P_{S(t)}; X_t])$ while freezing the pre-trained model parameters $\theta$.
Learning on each domain independently, we derive domain-specific prompts $P_{S1}, P_{S2}, ..., P_{S(\tau)}$, effectively condensing domain knowledge into a concise set of parameters. 

\input{sections/algorithm_icml}

\subsection{Temporal Prompt Learning} 
\label{method:temporal-prompt}

To capture concept drift over time, 
we employ a temporal prompt generator to encode the temporal dynamics into temporal prompts. This module takes in domain-specific prompts from source domains and produces future temporal prompts. Here, we utilize a single-layer transformer encoder module, denoted by $g_\omega$, as our temporal prompt generator. In order to incorporate information from the preceding domains, we apply sequential training. 
Starting from $t=2$, for each domain $D_t$ the temporal prompt generator $g_\omega$ receives domain-specific prompts, $P_{S1}, P_{S2}, \ldots , P_{S(t-1)}$, as input tokens. It then uses those prompts to generate the temporal prompts $P_{T2}, P_{T3}, \ldots , P_{T(t)}$. Specifically, as shown in Equation~\ref{eq:temporalpromp}, it generates the temporal prompt $P_{T(t)}  \in \mathbb{R}^n$ for domain $D_t$ from previous domain-specific prompts.
\begin{equation}
     P_{T(t)} = g_\omega (P_{S1:(t-1)}), \quad t = 2,\ldots,\tau
    \label{eq:temporalpromp}
\end{equation}

Moreover, to help capture generic information across all domains, we incorporate a learnable general prompt $P_G \in \mathbb{R}^n$.
The input $X_t$ is prepended by the generic prompt $P_G$ and the temporal prompt $P_{T(t)}$. 
The result, represented as $[P_{T(t)}; P_G; X_t]$, is fed into the frozen backbone network $f_\theta$. 
Both \( P_G \) and the temporal prompt generator $g_\omega$ are trained to maximize the likelihood $\mathcal{P}_\theta(Y_t | [P_{T(t)}; P_G;  X_t])$, while keeping the backbone network $f_\theta$ frozen. Temporal prompts $P_{T2}, P_{T3},\ldots, P_{T(\tau)}$ effectively capture temporal drift and help the pre-trained network to adapt to changes in the data distribution over time, and to anticipate future changes by capturing temporal trends.

\subsection{Inference}
During inference, the model utilizes the domain-specific prompts $P_{S1},P_{S2}, \ldots, P_{S(\tau)}$ and generates temporal prompts $P_{T2},P_{T3}, \ldots, P_{T(\tau+1)}$. To perform the target domain task, the frozen backbone receives the input $[P_{T(\tau+1)}; P_G; X_{(\tau+1)}]$ and predicts the output. 

A summary of prompt learning and inference is presented in \cref{alg:summary}.

%% file: sections/algorithm_icml.tex
\begin{algorithm*}
\caption{Prompt Learning and Inference}
\label{alg:summary}
\begin{algorithmic}[1]
\Require
    Source domains $\{D_t=(X_t, Y_t) | 1 \leq t \leq \tau\}$,
    Target domains $\{D_t=(X_t, Y_t) | t > \tau\}$,
    Pre-trained model to adapt $f_\theta$ parameterized by $\theta$,
    Temporal prompt generator $g_\omega$ parameterized by $\omega$
\Ensure
    Domain-specific prompts $P_{S1}, P_{S2}, \ldots, P_{S\tau}$,
    Temporal prompts $P_{T2}, P_{T3}, \ldots, P_{T(\tau+1)}$,
    Generic prompt $P_G$

\Procedure{DomainSpecificPromptLearning}{}
    \For{each domain $D_t$ in $\{D_t | 1 \leq t \leq \tau\}$}
        \State Prepend  $X_t$ with $P_{S(t)}$
        \State Process combined input $[P_{S(t)}; X_t]$ using frozen backbone $f_\theta$
        \State Train model to maximize likelihood $\mathcal{P}_{\theta}(Y_t | [P_{S(t)}; X_t])$ with $\theta$ fixed
    \EndFor
    \State Return domain-specific prompts $P_{S1}, P_{S2}, \ldots, P_{S\tau}$
\EndProcedure

\Procedure{TemporalPromptLearning}{}
    \State Initialize the temporal prompt generator $g_\omega$
    \For{each domain $D_t$ in $\{D_t | 2 \leq t \leq \tau\}$}
        \State Provide prompts $P_{S1}, P_{S2}, \ldots, P_{S(t-1)}$ to temporal prompt generator $g_\omega$
        \State Generate temporal prompt   $P_{T(t)}$
        \State Prepend input $X_t$ from domain $t$ with $P_G$ and $P_{T(t)}$
        \State Process input $[P_{T(t)}; P_G; X_t]$ using frozen backbone $f_\theta$
        \State Train model to maximize likelihood $\mathcal{P}_{\theta}(Y | [P_{T(t)}; P_G; X_t])$ with $\theta$ fixed
    \EndFor
\EndProcedure

\Procedure{Inference}{}
    \State Forecast $P_{T(\tau+1)}$ given domain-specific prompts $P_{S1}, P_{S2}, \ldots, P_{S\tau}$ and generic prompt $P_G$ from source domains 
    \State Predict $Y_{(\tau+1)}$ using the frozen backbone network $f_\theta$ and inputs $[P_{T(\tau+1)}; P_G; X_{(\tau+1)}]$
\EndProcedure

\end{algorithmic}
\end{algorithm*}

%% file: sections/experiments.tex
\section{Experiments}


\subsection{Baseline Methods} 

We compare our model with several state-of-the-art methods, including temporal domain generalization methods DRAIN~\cite{bai2023temporal} and GI~\citep{nasery2021training}, continuous domain adaption methods CDOT~\citep{ortiz2019cdot} and CIDA~\citep{wang2020continuously}, and prompting method ATTEMPT~\cite{asai2022attempt} to validate the effectiveness of our temporal prompts. The original DRAIN paper employed two fully connected layers (DRAIN-2FC) in both encoding and decoding functions to transform the latent representations between LSTM units. To potentially boost DRAIN's performance, we also explored using three and four linear layers in both encoding and decoding functions. We denote these models DRAIN-3FC and DRAIN-4FC, respectively. DRAIN-Best refers to the model achieving the highest performance using these configurations for the encoding/decoding functions.

We also compare against baseline methods that do not consider temporal drift, including {1) Vanilla-MLP, the MLP-based backbone network from DRAIN~\cite{bai2023temporal}, which is trained on the combined source domains, and 2) Vanilla-Transformer, our method's transformer-based backbone network, which is trained on the combined source domains.}

\subsection{Implementation Details}
\label{sec:implementation}

We utilize the Adam optimizer~\cite{kingma2014adam} and consistently set the learning rate to $1e-4$ across all datasets. Our system is implemented in PyTorch and runs on a workstation powered by a 2.10GHz Intel Xeon(R) Gold 6230 CPU with 20 cores, paired with an NVIDIA RTX 5000 GPU. For each dataset, we tune the hyperparameters based on the suggestions from~\cite{bai2023temporal}. Additional experiment settings and results (e.g., network architectures and additional ablation results) are provided in the appendix.

\begin{figure*}[tbh!]
\centering
\includegraphics[width=0.85\textwidth]{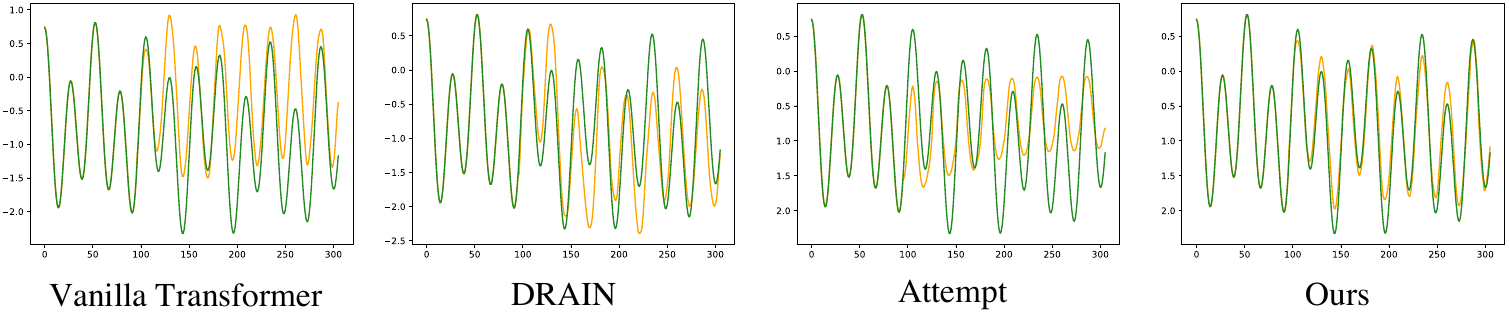}
\caption{Qualitative results on Sum of Cosines synthetic dataset generated with phase-frequency modification and addition of a variable cosine wave. }
\label{result:sys1-2}
\end{figure*}


\begin{figure}[tbh!]

\centering
\includegraphics[width=\linewidth]{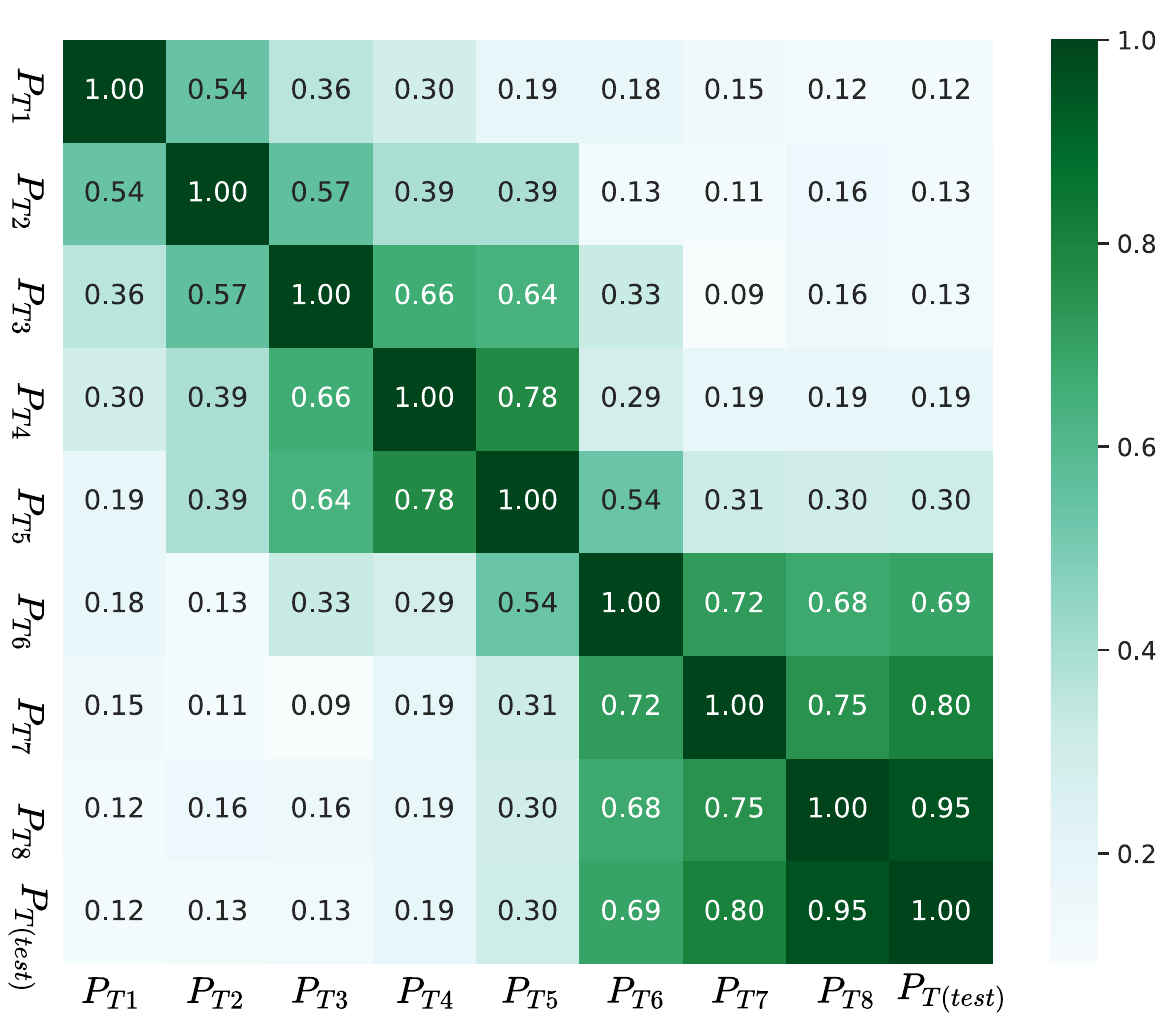}
\caption{
Pairwise comparisons of learned temporal prompts ($P_Ts$) across domains in the Sum of Cosines synthetic dataset with both types of synthetic drift. The values shown are cosine similarities.}
\label{result:sys:conf}
\end{figure}

\input{sections/syn_data}



\subsection{Main Experimental Results}


In this section, we demonstrate the generality of prompting-based temporal generalization using public datasets covering classification, regression, and time series forecasting tasks:
\begin{itemize}
\item Classification datasets: Rotated Moons (2-Moons)~\cite{nasery2021training}, Online News Popularity (ONP)~\cite{ben2010theory}, Electrical Demand (Elec2)~\cite{nasery2021training};
\item Regression datasets: House prices (House)~\cite{nasery2021training}, Appliances energy prediction (Appliance)~\cite{bai2023temporal};
\item Time series dataset: Crypto~\cite{crypto}.
\end{itemize}

For the classification and regression datasets (2-Moons, ONP, Elec2, House, and Appliance), we followed the procedure outlined in \cite{bai2023temporal} to partition the dataset into distinct temporal domains. The Crypto dataset contains eight features on historical trades (e.g., open and close prices) for 14 crypto currencies. Our goal is to generate 15-step predictions for the 15-minute relative future returns (i.e., the target), with each step representing a one-minute increment from the previous one. The data spans from 2018 until 2021 and we model each month as one domain. We used the initial $90\%$ of entries from each month in 2018, 2019, and 2020 for training (across 36 domains), while reserving the remaining $10\%$ of entries for \emph{in-domain} testing. The data from the first month of 2021 was designated for validation, with the subsequent three months of 2021 allocated for actual testing.

 Table~\ref{tab:Drain-compare} summarizes the classification and regression results in comparison to other state-of-the-art methods. The experiments are conducted ten times for each method on every dataset, with both the mean and the standard deviation reported. It is observed that our proposed method yields lower errors in all instances except for the 2-Moons dataset. 
 This may be due to the low dimensionality of the 2-Moons dataset (only 2 dimensions), which leads to less generalizable backbones for prompt-based approaches (as evidenced by the poor performance in ATTEMPT as well).

Table~\ref{tab:crypto} shows the time series forecasting results on the Crypto dataset. To ensure a fair comparison, DRAIN, ATTEMPT, and our method all adopt the same Vanilla-Transformer backbone network . We explored two settings: one with fixed-length input sequences where the look-back window is 15 steps  and the other with variable-length input sequences, where look-back window length varies from 7 to 15.
Our model is more accurate under both settings (with a lower RMSE) compared to DRAIN, Vanilla-Transformer, and ATTEMPT. Furthermore, our method is significantly more parameter and time efficient than the current state-of-the-art temporal domain generalization method, DRAIN. While ATTEMPT, also a prompt-based approach, matches our method's efficiency in terms of parameters and time, it falls short in performance due to its inability to model temporal drift. 


\input{tables/ablation_PT_PG}

\input{tables/ablation_domain_num}

\subsection{Ablation Studies and Additional Experiments}

We conducted ablation studies on the Crypto and Elec2 datasets to see the impact of the different types of proposed prompts. Table~\ref{tab:abl-promptsvarient} shows that both prompting mechanisms $P_T$ and $P_G$ contribute to better performance. Next, in order to study the impact of the number of training domains on method performance, we conducted an ablation study on Mackey-Glass synthetic data with varying numbers of training domains. ~\cref{tab:number_domain} shows that the performance of our method improves as the number of source domains increases, which is expected as a greater number of observed source domains make temporal patterns more evident. 

Finally, we experimented with a state-of-the-art multi-scale transformer architecture, Scaleformer~\cite{scaleformer}, on Mackey-Glass synthetic data. Table~\ref{tab:ad} shows the results, where Scaleformer has been applied on Informer~\cite{zhou2021informer}. The results confirm that our method generalizes to state-of-the-art transformers and can support further improvements by incorporating complementary advances in network design.

Further experiments on sequential versus non-sequential training paradigms, as well as the influence of hyperparameter configurations on model performance, can be found in the appendix.

  

\begin{table}[t]
    \centering
      \caption{Additional experimental results on Mackey-Glass with state-of-the-art transformers Scaleformer~\cite{scaleformer} and Informer~\cite{zhou2021informer}.}
    \begin{tabular}{c|c||c}
    \toprule
    \multirow{3}{*}{Method} &  \multicolumn{2}{c}{RMSE $\times 10^1$ $\downarrow$} \\ \cline{2-3}
    & Mackey-Glass & Mackey-Glass \\
    & \scriptsize{with $\sigma$-modification} & \scriptsize{with variable cosine} \\ \midrule
         \small{Vanilla-Trans. (VT)}  & 1.315 $\pm$0.07 &2.511$\pm$0.09\\
        \small{Informer} & 1.127 $\pm$0.05 & 2.074  $\pm$0.08  \\
       \small{Ours (on VT)}  & 0.982  $\pm$0.06  & 1.975$\pm$0.06   \\
        \small{Scaleformer} & 0.934 $\pm$0.00&  1.967 $\pm$0.01 \\
        \small{Ours (on Scaleformer)} & \bf {0.802 $\pm$0.01}&  \bf{1.852 $\pm$0.04}\\
        \bottomrule
    \end{tabular}
   
    \label{tab:ad}
\end{table}

%% file: sections/syn_data.tex
\input{tables/synt}

\subsection{Results on Synthetic Data}
\label{4.6}

To evaluate the proposed method under a controlled setting, we constructed synthetic datasets derived from the Mackey-Glass equations~\cite{mackey1977oscillation} and sums of cosines. For the former, we used 


\vspace{-0.7cm}
\begin{equation}
\begin{split}
    x(t+1)=x(t)+\beta \frac{x(t-\sigma)}{1+x(t-\sigma)^n}-\gamma x(t)
\label{eq-mackey}
\end{split}
\end{equation}

where $\beta=0.2, \gamma=0.1, n=15, \sigma=18, t_{\textit{max}}=2600$, and $x(t)=0.1 \textit{ if } t<18 $. For the latter, we used 

\begin{equation}
\begin{split}
    x(t) = \cos\left(a + \frac{\pi h}{\alpha}  t \right) +  \cos\left(b + \frac{\pi}{\beta} t \right)
\label{eq-cosine_wave}
\end{split}
\end{equation}

where $\alpha=100,  \beta =13, a =40, b =10,  h=1,$ and $0<t<2600 $.

We employed two strategies to induce temporal drift: parameter modification, and addition of a cosine wave with variable phases and frequencies across domains, given by 

\begin{equation}
     0.5 \times \cos\left( 100i + \frac{\pi(i+1)}{{300}} t \right)
     \label{cosine_add}
\end{equation}

for domain $i$.
For Mackey-Glass, we generated two datasets: one dataset modifying $\sigma=8+i\times 2$ for each domain $i$, 
and one dataset adding \cref{cosine_add} to \cref{eq-mackey}. For Sum of Cosines, we generated two datasets: one dataset modifying $a=i$ and $h=i+1$ for each domain $i$, and one dataset adding \cref{cosine_add} to the parameter-modified dataset. Visualizations of the synthetic datasets are shown in the appendix.




\input{tables/main_table}
\input{tables/crypto}

Results on the synthetic datasets are summarized in ~\cref{tab:synthetic_results}. We also qualitatively visualize the results on Sum of Cosines in ~\cref{result:sys1-2}. The proposed method consistently outperformed the Vanilla Transformer, DRAIN, and Attempt models on the synthetic data. Quantitatively, our model achieved the lowest MSE across both Mackey-Glass and Sum of Cosines datasets with either type of the temporal drifts. Qualitatively, it also demonstrated superior adaptability and accuracy. ~\cref{result:sys:conf} visualizes the pairwise cosine similarities among the learned temporal prompts across different domains for the Sum of Cosines dataset with both types of synthetic drift. It is observed that temporal prompts from neighboring domains have higher similarity than other domains. 

%% file: tables/synt.tex
\begin{table*}[ht]
\centering
\caption{Comparison of proposed method, vanilla transformer, and state-of-the-art approaches on synthetic datasets generated based on Mackey-Glass equations and Sum of Cosines, in mean square error $\times 10^1$.}
\scalebox{1}{\begin{tabular}{c|c|c||c|c}
\toprule
\multirow{4}{*}{{Method}} & \multicolumn{4}{c}{MSE $\downarrow$}\\ \cline{2-5}
& \multicolumn{2}{c||}{Mackey-Glass} & \multicolumn{2}{c}{Sum of Cosines} \\
 & \multirow{2}{*}{\small{with $\sigma$-modification}} & \multirow{2}{*}{\small{with variable cosine}} & \small{with phase-frequency}  & \small{with phase-frequency} \\
 & & & \small{modification} & \small{modification + var. cosine} \\ 
\midrule 
DRAIN-Best & 1.140 $\pm$0.08 & 2.164 $\pm$ 0.08 & 0.085 $\pm$0.01 & 2.93  $\pm$0.07 \\
Vanilla Transformer & 1.315 $\pm 0.07$ & 2.511 $\pm$0.09 & 0.1191 $\pm$0.19 & 3.708 $\pm$0.07 \\
Attempt & 1.278 $\pm$0.10 & 2.199 $\pm$0.12 & 0.091 $\pm$0.02 & 2.974 $\pm$0.10 \\
Ours & \bf{0.982 $\pm$0.06} & \bf{1.975 $\pm$0.05} & \bf{0.068$\pm$0.00} & \bf{2.489$\pm$0.07} \\
\bottomrule
\end{tabular}}
\label{tab:synthetic_results}
\end{table*}

%% file: tables/main_table.tex
\begin{table*}[ht!]

\centering

\caption{Performance comparison of all methods in terms of classification error (in \%) for classification tasks and mean absolute error (MAE) for regression tasks (both smaller the better.) Results of comparison methods on all datasets are reported from~\cite{bai2023temporal}. ``-'' denotes that the method could not converge on the specific dataset.}

\scalebox{1}{\begin{tabular}{c|c|c|c||c|c}
\toprule
\multirow{2}{*}{{Method}} & \multicolumn{3}{c||}{{Classification} [\% error $\downarrow$]} & \multicolumn{2}{c}{{Regression} [MAE $\downarrow$]}\\
 & 2-Moons& ONP& Elec2 &House &Appliance \\ 
\midrule

Vanilla-MLP & 22.4 $\pm$ 4.6&33.8 $\pm$ 0.6 &23.0 $\pm$ 3.1 & 11.0 $\pm$ 0.36 &10.2 $\pm$ 1.1 \\
CDOT~\citep{ortiz2019cdot} & 9.3 $\pm$ 1.0&34.1 $\pm$ 0.0 & 17.8 $\pm$ 0.6& -&- \\
CIDA~\citep{wang2020continuously} & 10.8 $\pm$ 1.6 &34.7 $\pm$ 0.6& 14.1 $\pm$ 0.2 & 9.7 $\pm$ 0.06 &8.7 $\pm$ 0.2\\
GI~\citep{nasery2021training}& 3.5 $\pm$ 1.4&36.4 $\pm$ 0.8& 16.9 $\pm$ 0.7 & 9.6 $\pm$ 0.02 & 8.2 $\pm$0.6\\
DRAIN~\cite{bai2023temporal} & \bf{3.2 $\pm$ 1.2} & 38.3 $\pm$ 1.2 & 12.7 $\pm$ 0.8& 9.3 $\pm$ 0.14&6.4 $\pm$ 0.4\\
Vanilla-Transformer& 25.2 $\pm$ 0.9 & 33.6 $\pm$ 0.5 &22.5 $\pm$ 0.6&11.8$\pm$ 0.3 &5.6 $\pm$ 0.4 \\
Attempt~\cite{asai2022attempt} & 21.1 $\pm$ 1.1& 34.1 $\pm$0.6 &12.3 $\pm$0.8 &9.0 $\pm$0.4 &4.9 $\pm$0.5\\
Ours &8.1 $\pm$ 1.0 & \bf{32.7 $\pm$ 0.7 } & \bf{10.6$\pm$ 0.9}&\bf{8.9$\pm$ 0.20} &\bf{4.7 $\pm$ 0.3}\\

\bottomrule
\end{tabular}}
\label{tab:Drain-compare}
\end{table*}

%% file: tables/crypto.tex
\begin{table*}[tbh!]
\caption{{Performance comparison of our method with DRAIN~\cite{bai2023temporal} and ATTEMPT~\cite{asai2022attempt} on Crypto dataset in terms of root mean square error $\times 10^3$.}}
\centering
\scalebox{1}{
\begin{tabular}{l|l|c|c|c|c|c|c}
\toprule
len. & Method & \#Parameter & Training time (s) &In domain&$D_{t1}$ &$D_{t2}$ &$D_{t3}$ \\ \midrule
 \multirow{7}{*}{\rotatebox{90}{Fixed} }& DRAIN-2FC & 8M &1634& 3.96 $\pm$0.07 & 4.27 $\pm$0.08&7.03$\pm$0.1& 7.24 $\pm$0.09   \\ 

&DRAIN-3FC &239M &2520&3.82 $\pm$0.09  & 3.90$\pm$ 0.08  &6.75 $\pm$0.08  & 6.89  $\pm$0.10 \\ 

& DRAIN-4FC &254M &2827& 3.60$\pm$0.10  & 3.61$\pm$0.09   &6.69  $\pm$ 0.12  & 6.70 $\pm$0.14 \\
\cmidrule{2-8} 

& Vanilla-Trans.&69K &239 & 4.00$\pm$ 0.05 &4.42$\pm$0.03& 7.19 $\pm$0.06 & 7.43 $\pm$0.06\\

& Attempt & 93K & 684& 3.57$\pm$0.06 & 4.03$\pm$0.08  & 7.22 $\pm$0.14&7.45 $\pm$0.16 \\
& Attempt-m & 93K & 684&3.54 $\pm$0.10  & 3.79$\pm$0.11  & 6.96 $\pm$0.15 &7.35 $\pm$ 0.15\\

 & Ours & 94K & 717 & \bf{3.44 $\pm$0.06} & \bf{3.53$\pm$0.06 } &\bf{6.61 $\pm$0.05}& \bf{6.74 $\pm$0.08 }\\

  \cmidrule{1-8}
 \multirow{7}{*}{\rotatebox{90}{Not-Fixed} }& DRAIN-2FC &8M& 1634 &4.97$\pm$ 0.08 & 5.22$\pm$ 0.08& 7.78$\pm$ 0.12  &7.98 $\pm$ 0.11 \\

& DRAIN-3FC &239M &2520&4.61 $\pm$ 0.09& 4.95 $\pm$0.07 &7.38 $\pm$ 0.09  &7.47 $\pm$ 0.13\\

 & DRAIN-4FC &254M & 2827 & 3.66$\pm$ 0.08 & 3.74$\pm$ 0.08 &6.82 $\pm$ 0.10& 7.03 $\pm$ 0.15 \\
\cmidrule{2-8}
 &Vanilla-Trans.&69k &239& 4.08 $\pm$ 0.08 & 4.44$\pm$ 0.07& 7.28 $\pm$ 0.09 &7.55 $\pm$ 0.08\\
& Attempt & 93K & 684&3.85$\pm$0.10 &4.29$\pm$ 0.11 & 7.51 $\pm$ 0.12 &7.75 $\pm$0.13  \\\ 
& Attempt-m & 93K & 684& 3.79$\pm$ 0.12 & 4.12$\pm$ 0.10& 7.16$\pm$ 0.17 &7.43 $\pm$ 0.15\\

 &Ours & 94K & 717 & \bf{ 3.53$\pm$ 0.06 }& \bf{3.57$\pm$ 0.07} & 
 \bf {6.66 $\pm$0.10} & \bf{6.89 $\pm$ 0.09}\\
 
\bottomrule
\end{tabular}}

\label{tab:crypto}
\end{table*}

%% file: tables/ablation_PT_PG.tex
\begin{table}[t]
 \centering
\captionof{table}{Ablation of effect of $P_G,P_T$ using Crypto and Elec2 datasets. \checkmark indicates the prompt is used.}
\begin{tabular}{c c |c|c|c||c}
\toprule
&& \multicolumn{3}{c||}{Crypto [RMSE $\times 10^3$ $\downarrow$]}  & \multicolumn{1}{c}{Elec2} [MAE  $\downarrow$]\\
$P_G$ & $P_T$ & $D_{t1}$ & $D_{t2}$ & $D_{t3}$ & $D_t$ \\ \midrule
\checkmark & & 3.57 & 6.66&6.84 &14.9 \\ 
 &\checkmark& 3.53 & 6.71& 6.80& 14.7\\ 
\checkmark & \checkmark& 3.53 &6.61 &6.74 &10.6\\\bottomrule
\end{tabular}
\label{tab:abl-promptsvarient}
\end{table}

%% file: tables/ablation_domain_num.tex
\begin{table}[t]
    \centering
      \caption{Impact of number of training domains on vanilla transformer and our method.}
    \begin{tabular}{c|c|c||c|c}
    \toprule
    &  \multicolumn{4}{c}{MSE $\times 10^1 $ $\downarrow$} \\ \cline{2-5}
   \# Training & \multicolumn{2}{c||}{Mackey-Glass} & \multicolumn{2}{c}{Mackey-Glass} \\
   domains & \multicolumn{2}{c||}{\small{with $\sigma$-modification}} & \multicolumn{2}{c}{\small{with variable cosine}} \\
   &  Vanilla- & Ours & Vanilla- & Ours  \\ 
   & Trans. & & Trans. & \\ \midrule
       
        4 &1.818   & 1.305 & 3.007 &2.581 \\
        9 & 1.315  & 0.982 &  2.511  &1.975  \\
        19 & 0.877& 0.787& 3.326  & 2.547 \\
        49 & 0.930& 0.739 & 1.645 &  1.440\\
        \bottomrule
    \end{tabular}
  
    \label{tab:number_domain}
\end{table}

%% file: sections/conclusion_icml.tex
\section{Conclusion}

Machine learning models deployed in real-world settings face many operational challenges, not the least of which is the need to maintain stable and robust performance long after initial release. Data distribution shifts over time, such as evolving customer preferences, user behavior, or economic conditions, underscore the need to adapt machine learning models deployed in the real world. We believe that a small yet important part of the solution lies in anticipating and adapting to temporal changes in the data. Enabling a trained model to generalize to unseen future time steps requires principled ways not only to learn domain-invariant features, but also to capture the temporal dynamics of the data. In this paper, we described a general and efficient method for temporal domain generalization using learnable real-valued prompts. We hope that our research inspires further work in ensuring that machine learning models remain reliable over time.

\paragraph{ICML impact statement.} Our research addresses the problem of adapting trained machine learning models to data distribution changes over time. To our knowledge, it does not introduce new ethical or societal risks that need to be specifically highlighted here.

%% file: sections/appendix.tex
\onecolumn
\section{Appendix}

\subsection{Network architectures and experimentation details} Below, we detail the architecture and other specific experiment details for each dataset.

\mysubsubsection{Architecture of frozen backbone network} We choose backbones for each dataset to enable a fair comparison with state-of-the-art methods.

For the time series dataset {\bf Crypto}, the initial inputs are passed through a linear layer, resulting in 64-dimensional embeddings. These embeddings are then processed by a transformer encoder layer. The transformer comprises a single encoder layer with four heads, and the hidden layers with dimensionality of 128. Finally, the output is passed through another linear layer to achieve the desired output size. We utilize the mean squared error (MSE) loss for Crypto dataset.

For the datasets that are reported 
in DRAIN~\citep{bai2023temporal}, the initial inputs for {\bf Elec2}, {\bf 2Moons}, {\bf House}, and {\bf Appliance} are transformed through a linear layer to produce 128-dimensional embeddings, whereas for {\bf ONP} it is a 32-dimensional embedding. These embeddings are subsequently processed by a transformer encoder layer. Notably, to align closely with the DRAIN paper's structure, our transformer encoder employs just one linear layer in the feed-forward segment, as opposed to the conventional two. The transformer setup involves a single encoder layer with one head. The hidden layers maintain a 128-dimensional structure for all datasets, with the exception of {\bf ONP}, which is set at 64. Outputs are then channeled through another linear layer to derive the desired size. For regression datasets, we adopt the mean squared error (MSE) loss, and for classification datasets, we use binary cross-entropy loss.

\mysubsubsection{Domain-specific prompts} Domain-specific prompts are learnable parameters, whose sizes match the embedding dimensions for each dataset.

\mysubsubsection{{Temporal prompt} generator} We employ a transformer with a single encoder layer and 1 heads as our temporal prompt generator. The transformer's hidden layers have a consistent 128-dimensional configuration.

\subsection{Non-sequential temporal prompt learning}
\label{non-seq}
In the main paper, temporal prompts are generated sequentially. An alternative option is to generate them non-sequentially. In this experiment, we opt for a non-sequential training paradigm, wherein the model is exposed to all source domains simultaneously during the training process. To be precise, the temporal prompt generator, denoted as $g_{\omega}$, takes all domain-specific prompts $P_{S1}, P_{S2},\ldots, P_{S(\tau)}$, and generates temporal prompts $P_{T2}, P_{T3},\ldots, P_{T(\tau+1)}$. Table~\ref{tab:appendix1} compares performance of sequential temporal prompt generation vs non-sequential prompt generation, and it can be seen that performance is on par with the main method. 




\begin{table}[tbh]

 \centering
   \caption{Comparing sequential temporal prompt generation vs non-sequential one. }
    \begin{tabular}{c|c|c|c||c|c}
    \toprule
    \multirow{2}{*}{{Method}} & \multicolumn{3}{c||}{{Classification error [in \% $\downarrow$ ]}} & \multicolumn{2}{c}{{Regression [MSE  $\downarrow$]}}\\  
           & 2-Moons    &     ONP  &   Elec2 &  House &  Appliance \\ 
    \midrule 
Vanilla-Transformer& 25.2 $\pm$ 0.9 & 33.6 $\pm$ 0.5 &22.5 $\pm$ 0.6&11.8$\pm$ 0.3 &5.6 $\pm$ 0.4 \\
Attempt~\cite{asai2022attempt} & 21.15 $\pm$ 1.1& 34.10 $\pm$0.6 &12.26 $\pm$0.8 &9.0 $\pm$0.4 &4.9 $\pm$0.5\\
Ours &\bf{8.1 $\pm$ 1.0} & 32.7 $\pm$ 0.7  & \bf{10.6$\pm$ 0.9}&8.9$\pm$ 0.20 &\bf{4.7 $\pm$ 0.3}\\

      Ours (not sequential) &8.4 $\pm$ 0.9 & \bf{31.8 $\pm$ 0.7}  &11.2$\pm$ 0.8  &\bf{ 8.6$\pm$ 0.14} &4.9 $\pm$ 0.4  \\

    \bottomrule
    \end{tabular}%
  
    \label{tab:appendix1}
\end{table}

\subsection{Impact of embedding and prompt size on model performance }


  

Table~\ref{tab:abl-promptsize} shows ablations on embedding and prompt size. {It is observed that for Crypto dataset, embedding/prompting size 64 and 128 provide similar better performance, and smaller embedding/prompting size results in a more parameter-efficient network; 64 is selected for better model size and performance tradeoff.}

\begin{table}[tbh!]
 
\centering
 \captionof{table}{Impact of prompt size and embedding size using Crypto dataset,  in terms of root mean square error $\times 10$.}

\scalebox{1} {\begin{tabular}{c |c|c|c||c|c|c}
\toprule
 \multirow{1}{*}{ Prompt size \& } & \multicolumn{3}{c||}{{Vanilla Transformer }}&
\multicolumn{3}{c}{Temporal prompting} \\
  Embedding size &$D_{t1}$&$D_{t2}$ & $D_{t3}$ &$D_{t1}$&$D_{t2}$ & $D_{t3}$ \\ \midrule

32 &4.20 & 7.20 &7.45 &3.57 &6.64 &6.85 \\
64 & 4.42 & 7.19 &7.43 &3.53 &6.61 &6.74 \\
128 &4.52 & 7.59 &7.79 &3.45 &6.58 &6.79 \\
256 &4.45 & 7.25 &7.39 &3.45 &6.64 &6.79\\ \bottomrule

\end{tabular}}

\label{tab:abl-promptsize}
\end{table}

\subsection{Impact of temporal prompting module layers on model performance }
Table~\ref{tab:number_layers} presents an additional study on the effect of the number of layers in the temporal prompt generation module using the Mackey-Glass data.

\begin{table}[tbh!]
    \centering
      \caption{Impact of temporal prompting module layers on model performance in terms of MSE.}
    \begin{tabular}{c|c|c}
    \toprule
       \multirow{2}{*}{Number of Layers}  &  Mackey-Glass & Mackey-Glass \\
       & with $\sigma$-modification &   with variable cosine \\ \midrule

        Vanilla Transformer (0) &0.1315 &0.2511 \\
        1 & 0.0982&0.1975  \\
        2 & 0.0950  & 0.2053 \\
        3 &0.1022& 0.2119 \\
        
        \bottomrule
    \end{tabular}
  
    \label{tab:number_layers}
\end{table}

\newpage
\subsection{Visualization of synthetic data }

In this section we provide visualizations of the synthetic data from section~\ref{4.6}. 
\begin{figure}[h!]
\centering
\includegraphics[width=0.95\textwidth]{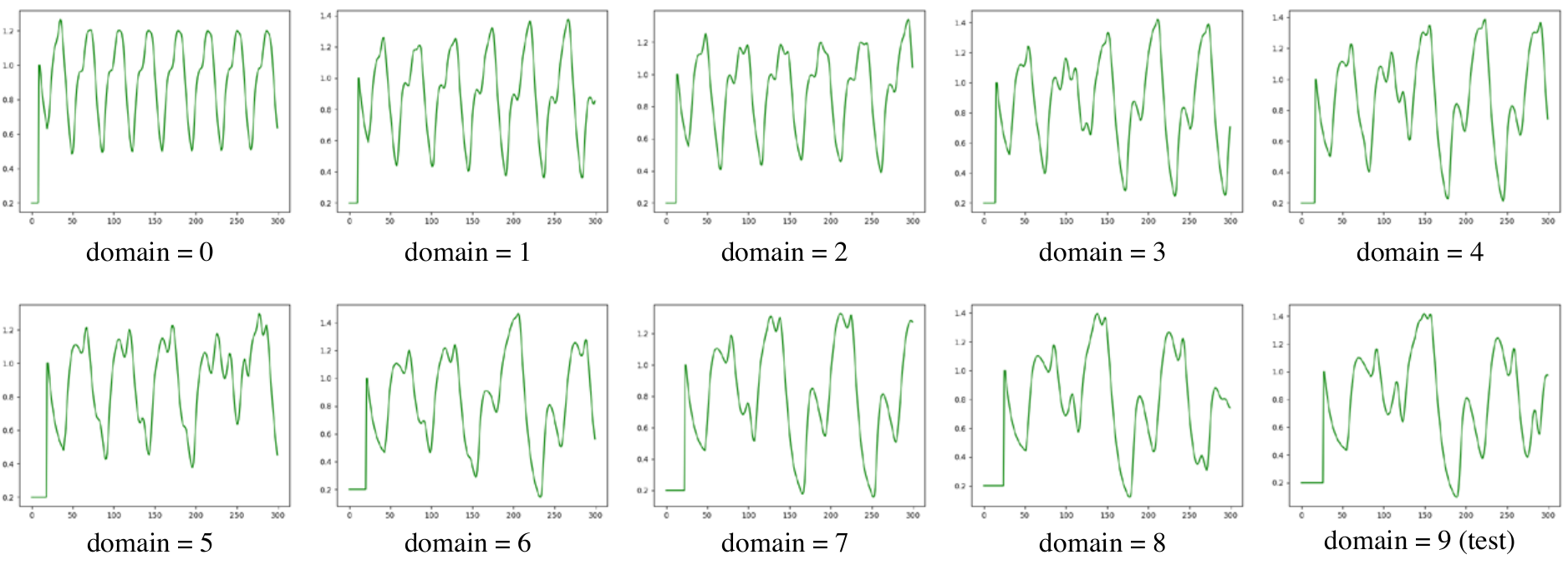}
\caption{Applying temporal shift to Mackey-Glass time series by modifying $ \sigma $. }
\label{sys1-1}
\end{figure}
\begin{figure}[tbh!]
\centering
\includegraphics[width=0.95\textwidth]{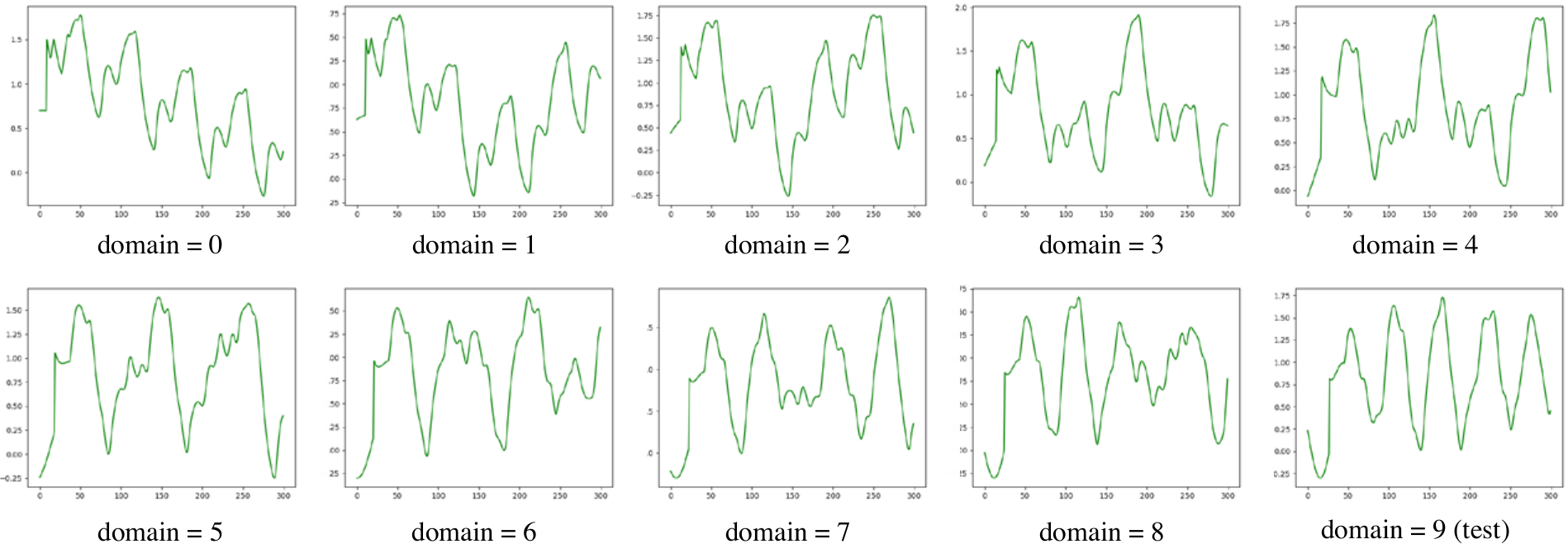}
\caption{Applying temporal shift to Mackey-Glass time series by adding variable cosine wave. }
\label{sys1-2}
\end{figure}

\begin{figure}[tbh!]
\centering
\includegraphics[width=0.95\textwidth]{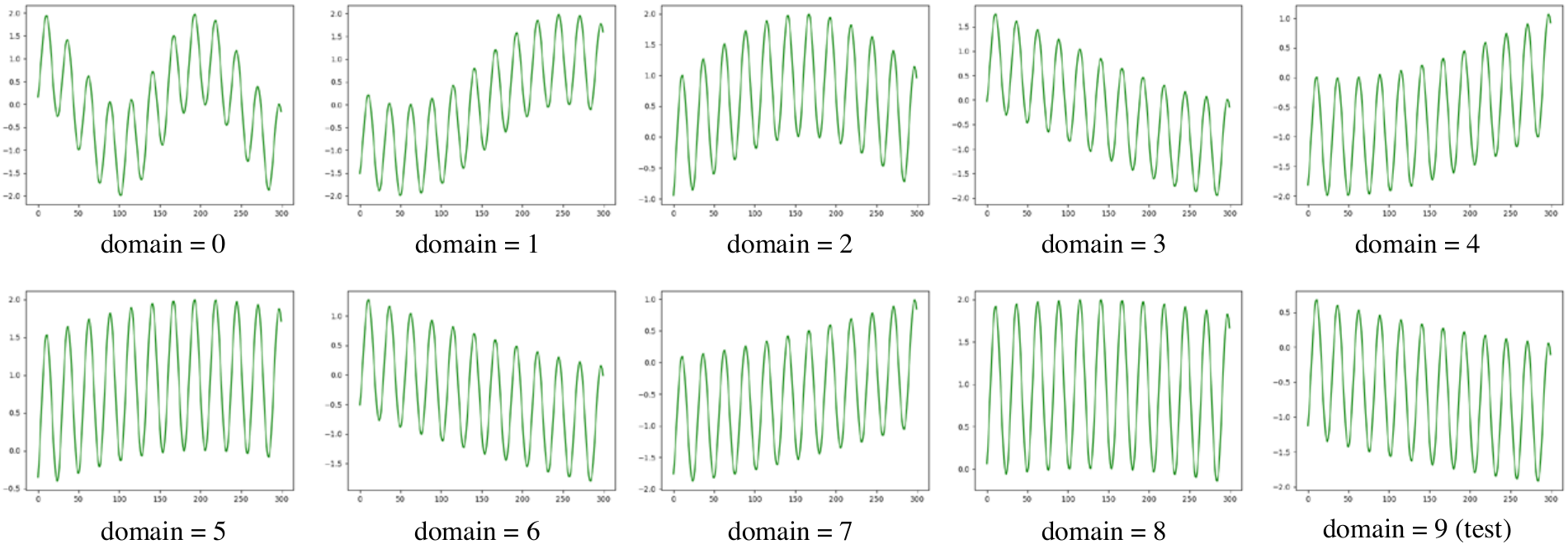}
\caption{Applying temporal shift to Sum of Cosines time series by modifying phase and frequency. }
\label{sys2-1}
\end{figure}

\begin{figure}[tbh!]
\centering
\includegraphics[width=0.95\textwidth]{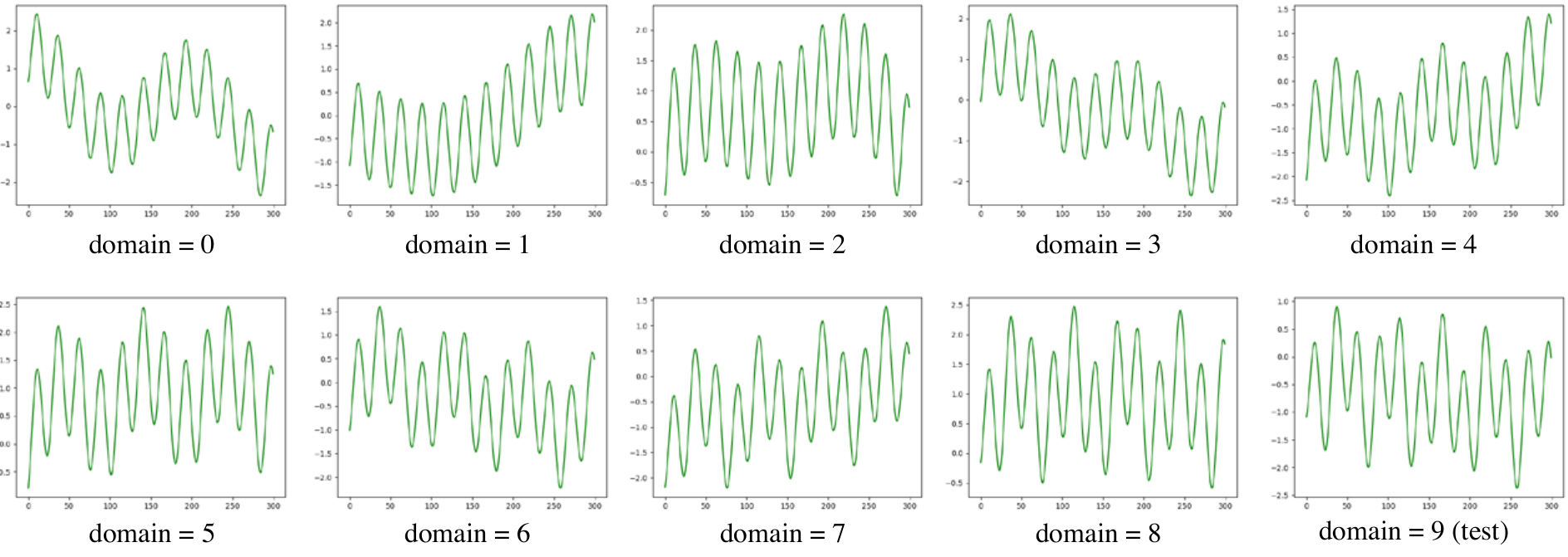}
\caption{Applying temporal shift to Sum of Cosines time series by modifying phase and frequency, and adding another variable cosine wave. }
\label{data:sys2-2}
\end{figure}